\documentclass[10pt,twocolumn,letterpaper]{article}

\usepackage{iccv}
\usepackage{times}
\usepackage{epsfig}
\usepackage{graphicx}
\usepackage{amsmath}
\usepackage{amssymb}

\usepackage{subfigure}
\usepackage[noend]{algpseudocode}
\usepackage{algorithmicx,algorithm}
\usepackage[pagebackref=true,breaklinks=true,colorlinks,bookmarks=false]{hyperref}

\iccvfinalcopy 

\def\iccvPaperID{2615} 

\ificcvfinal\fi

\begin{document}

\title{Interpolated Convolutional Networks for 3D Point Cloud Understanding}
\author{Jiageng Mao\quad
	Xiaogang Wang\quad
	Hongsheng Li\\
	CUHK-SenseTime Joint Laboratory, The Chinese University of Hong Kong\\
	{\tt\small \{maojiageng@link, xgwang@ee, hsli@ee\}.cuhk.edu.hk}
}
\maketitle
\ificcvfinal\thispagestyle{empty}\fi

\begin{abstract}
	Point cloud is an important type of 3D representation. However, directly applying convolutions on point clouds is challenging due to the sparse, irregular and unordered data structure. In this paper, we propose a novel Interpolated Convolution operation, InterpConv, to tackle the point cloud  feature learning and understanding problem. The key idea is to utilize a set of discrete kernel weights and interpolate point features to neighboring kernel-weight coordinates by an interpolation function for convolution. A normalization term is introduced to handle neighborhoods of different sparsity levels. Our InterpConv is shown to be permutation and sparsity invariant, and can directly handle irregular inputs. We further design Interpolated Convolutional Neural Networks (InterpCNNs) based on InterpConv layers to handle point cloud recognition tasks including shape classification, object part segmentation and indoor scene semantic parsing. Experiments show that the networks can capture both fine-grained local structures and global shape context information effectively. The proposed approach achieves state-of-the-art performance on public benchmarks including ModelNet40, ShapeNet Parts and S3DIS.
\end{abstract}

\section{Introduction}
Point cloud is an important data format obtained by 3D sensors and has shown extensive usage in many real-world tasks including autonomous driving~\cite{chen2017multi}, robotics~\cite{rusu2008towards}, etc. Efficient learning from point cloud data remains a challenge to the research community, given the fact that point clouds are usually irregular, unordered and sparse.

\begin{figure} 
	\centering 
	\subfigure[3D convolutions with rasterization]{\label{fig1.1}
		\includegraphics[width=0.45\linewidth]{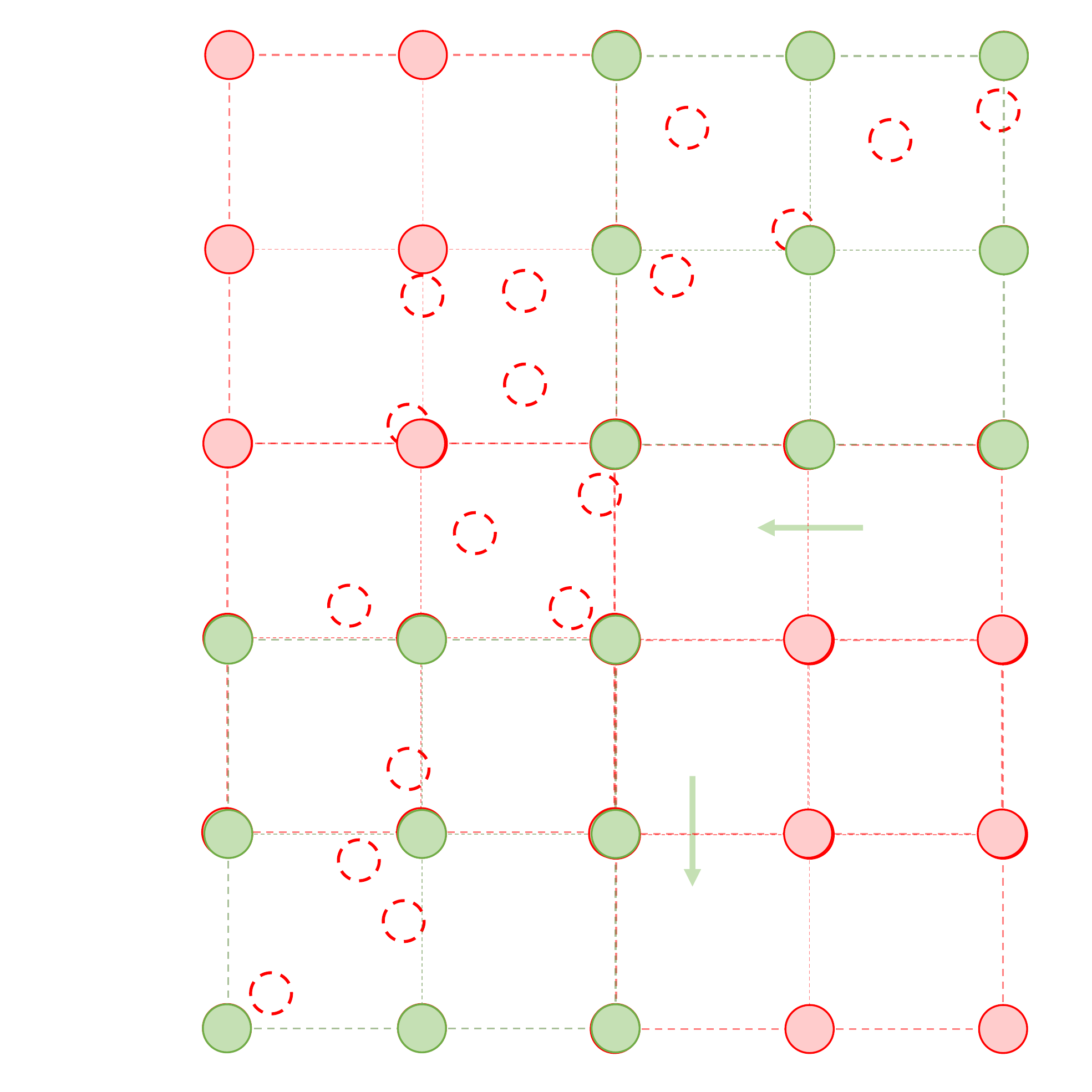}}
	\hspace{0.01\linewidth}
	\subfigure[Graph neural networks]{\label{fig1.2}		
		\includegraphics[width=0.45\linewidth]{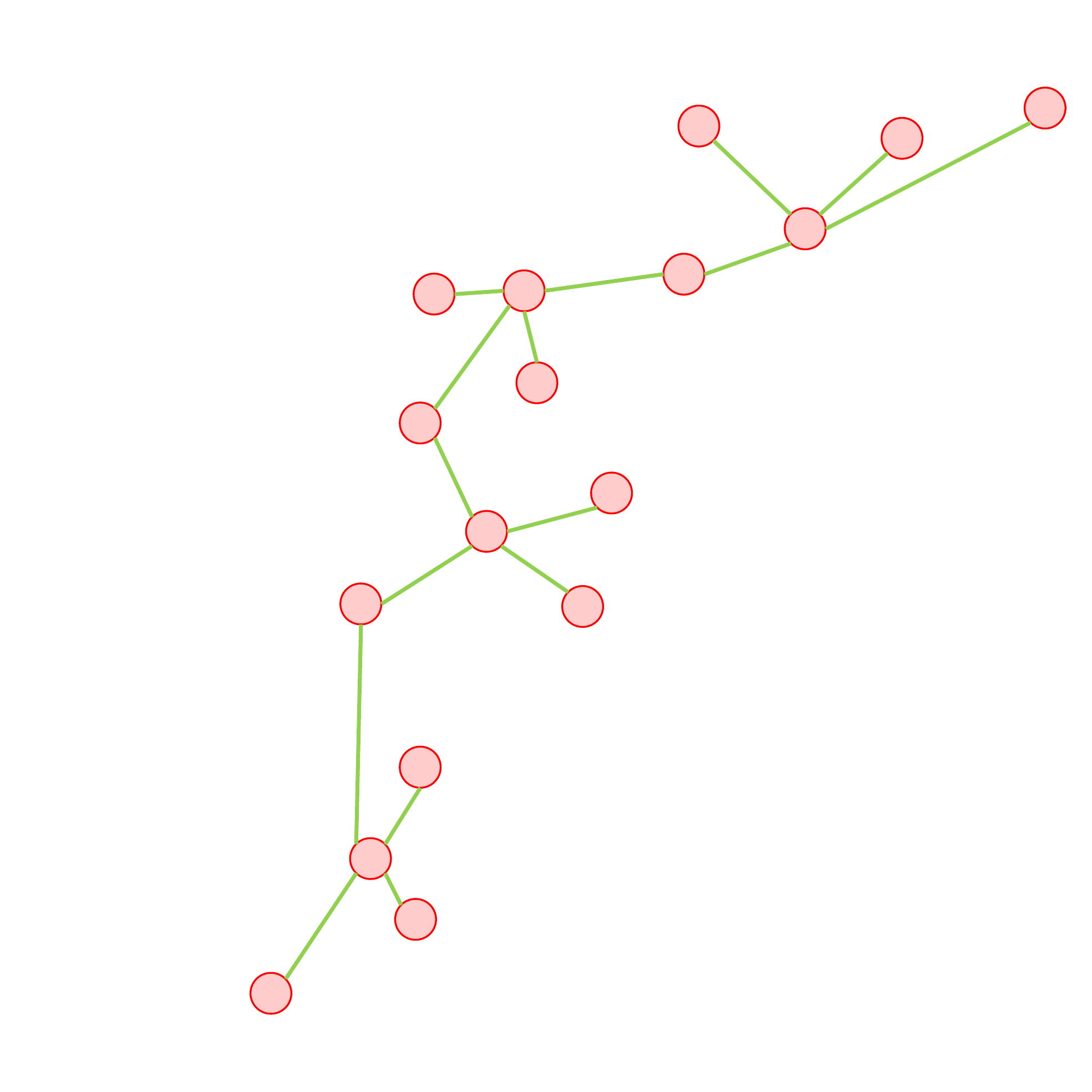}}	
	\vfill	
	\subfigure[InterpConv with trilinear interpolation]{\label{fig1.3}
		\includegraphics[width=0.45\linewidth]{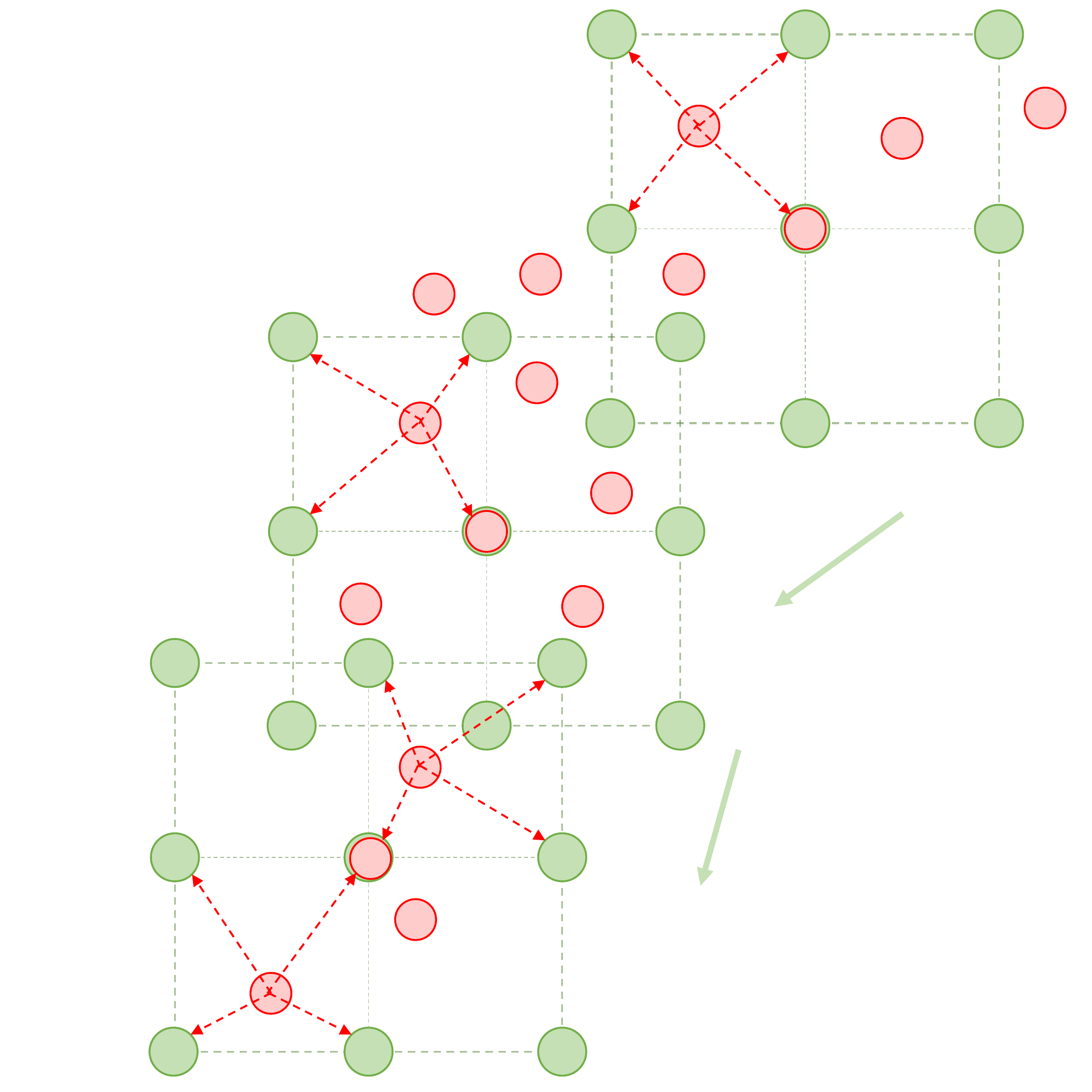}}
	\hspace{0.01\linewidth}	
	\subfigure[InterpConv with Gaussian interpolation]{\label{fig1.4}		
		\includegraphics[width=0.45\linewidth]{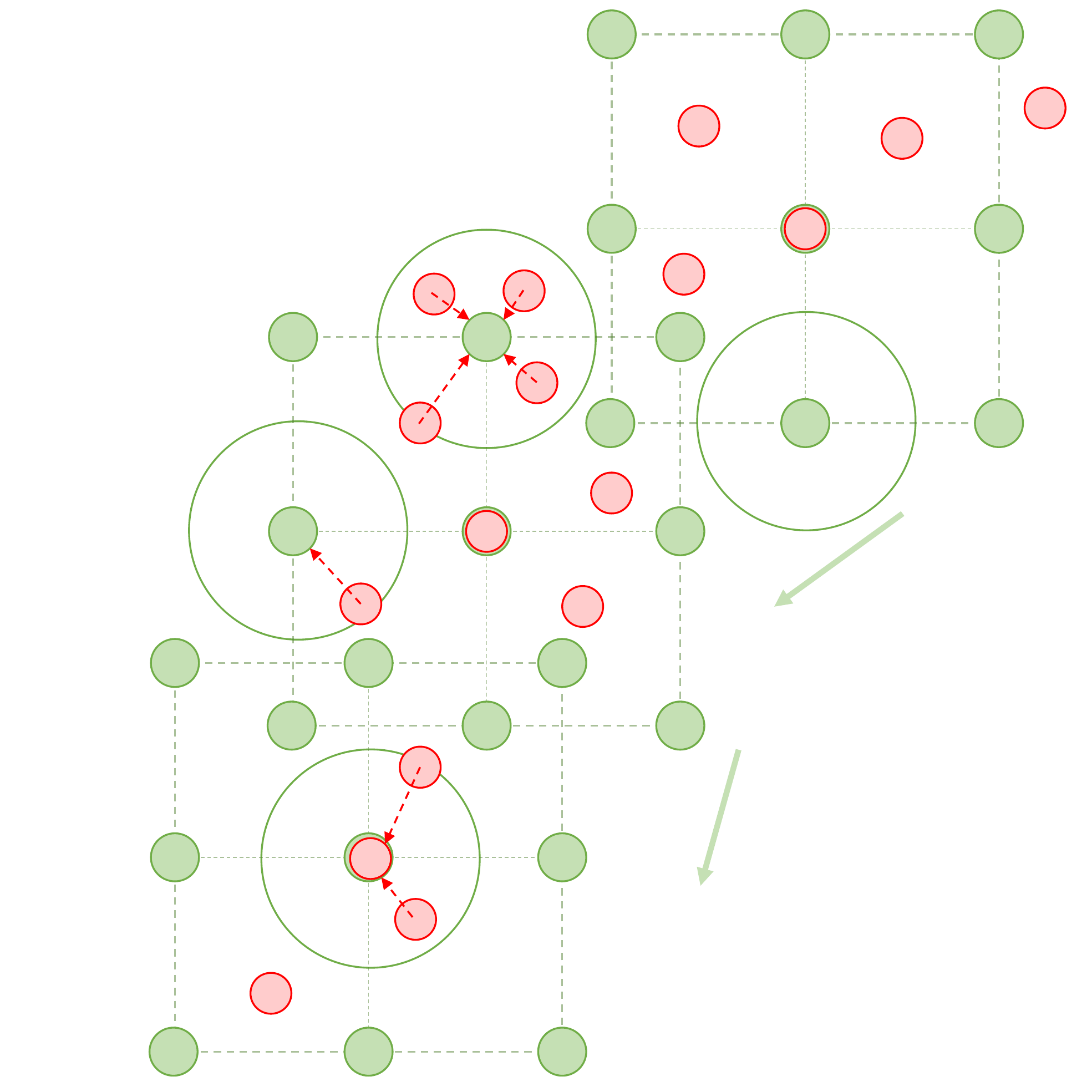}}	
	\caption{Illustration of different types of convolutions on point clouds. Red points denote point clouds and green points denote spatially-discrete kernel weights. Green lines in (b) denote continuous convolutional kernels. Our InterpConv directly takes irregular point clouds as inputs and interpolates point features to the neighboring kernel weights by an interpolation function.}	
	\label{fig1}
\end{figure}

In view of the great success of convolutional neural networks (CNNs) on 2D images, many endeavors have been made to adapt the convolution operation to 3D point clouds. Currently there are two main approaches to tackle this problem. The first type of attempts~\cite{maturana2015voxnet} is to directly rasterize irregular point clouds into regular voxel grids, and adopts standard 3D convolutions to learn shape features. However, the transformation of irregular inputs leads to a loss of geometric information, and convolutions on dense voxel grids lead to heavy computational burden. Other approaches~\cite{wang2018deep,wang2018dynamic,qi20173d,xu2018spidercnn,shen2018mining,li2018pointcnn,wu2018pointconv,tatarchenko2018tangent} build local graphs in the neighborhood of each point in the Euclidean or feature space, then a continuous convolutional kernel is applied on each edge of the graph to learn geometric features. Continuous kernels are commonly modeled by Multi-layer Perceptrons (MLPs). These graph-based methods are able to directly process irregular data structure but have some drawbacks. The construction of local graphs is not sparsity invariant. Namely, different point cloud densities sampled from the same object surface lead to different neighborhood selections, thus it may produce various graph construction results. Besides, compared with discrete convolutions, utilizing MLPs to learn an arbitrary continuous function does not work well in practice~\cite{xu2018spidercnn}. 

In this paper, we propose a novel Interpolated Convolution operation (InterpConv) to address the existing problems in graph and 3D convolutional neural networks. Key to our approach is the use of discrete convolutional kernels and an interpolation function to explicitly measure geometric relations between input point clouds and kernel-weight coordinates. Unlike 3D convolutions which have to transform inputs into regular grids, our InterpConv directly takes irregular point clouds as inputs. Each $n \times n \times c$ convolutional kernel is split into $n^{2}$ kernel weights, each of which has a $1 \times c$ weight vector and its own coordinate $p^{\prime}$ relative to the kernel center. The center of discrete convolutional kernels can be placed at any location in the 3D space and then the kernel-weight absolute coordinates can be determined for each kernel. Input points are interpolated to neighboring kernel-weight coordinates by an interpolation function. To guarantee InterpConv to be sparsity invariant, normalization on points is adopted in the neighborhood of each kernel weight vector. Finally, weighted convolutions can be calculated between kernel weight vectors and point cloud features that are associated to them. With spatially-discrete convolutional kernel weights and an explicitly defined interpolation function, our approach performs better than graph-based methods, which use continuous functions as convolutional kernels and implicitly learn geometric relations. See Figure \ref{fig1} for illustration.

We further propose Interpolated Convolutional Neural Networks (InterpCNNs) based on InterpConvs. The classification network is composed of multi-layer and multi-receptive-field InterpConv blocks, which can capture both fine-grained geometric structures and context information. The segmentation network explores a deeper architecture to predict semantic labels for all input points. We evaluate our networks on several benchmark datasets, including ModelNet40~\cite{chang2015shapenet}, ShapeNet Parts~\cite{yi2016scalable} and S3DIS~\cite{armeni20163d}. Experiments show that our approach achieves state-of-the-art performances on those datasets. 

The key contributions of our work are as follows:
\begin{itemize}
	\setlength{\itemsep}{0pt}
	\item We propose a novel Interpolated Convolution operation (InterpConv) to effectively deal with point cloud recognition problems. Such an operation is permutation and sparsity invariant, and can directly handle irregular point clouds;
	\item We design Interpolated Convolutional Neural Networks based on InterpConvs. The networks perform better than Graph Neural Networks (GNNs) and 3D Convolutional Neural Networks (3D ConvNets) on point cloud recognition and segmentation problems.
\end{itemize}

\section{Related Work}
Our approach is closely related to other deep learning methods on point clouds. We introduce literatures on point cloud feature learning by regular grids and irregular inputs.

\textbf{Learning from point clouds by regular grids.} When facing irregular point clouds as inputs, an intuitive way is to transform this irregular data structure into regular grids. Some approaches~\cite{kanezaki2018rotationnet,su2015multi,wang2017dominant} transform 3D objects or point clouds into 2D regular grids, namely, images by multi-view projection, then 2D CNNs are utilized to learn from these images. Those methods work well owing to the great success of 2D CNNs on images. However, not all the geometric information is kept during projection, and those approaches are usually inefficient and time-consuming when handling sparse point cloud data.

An alternative approach is to rasterize point clouds into 3D regular grids. VoxNet~\cite{maturana2015voxnet} transforms original point cloud data into occupancy grids, which store binary values to indicate whether the spaces are occupied. Then a 3D CNN is applied to learn from these voxel grids. The rasterization process loses some fine-grained geometric features, and 3D convolutions are both time and memory consuming. OctNet~\cite{riegler2017octnet} exploits the sparsity of voxel grids and uses unbalanced octrees to hierarchically partition the space, which saves much memory. Some other efforts~\cite{li2016fpnn,wang2015voting,rethage2018fully} have also been made to ease the computational burden but still cannot solve the loss of geometric information during rasterization. 

Compared with the above-mentioned approaches, our approach directly takes irregular point clouds as inputs without rasterization, which is time-saving and accurate.  

\textbf{Learning from point clouds by irregular inputs.} Recently there are many works trying to directly process irregular point cloud data. Pioneering work PointNet~\cite{qi2017pointnet} utilizes shared MLPs and a maxpooling layer, which is permutation invariant, to tackle unordered inputs and learn a global representation. PointNet++~\cite{qi2017pointnet++} exploits local structures by grouping and sampling point clouds, then a PointNet is applied in each group to aggregate local features. However, how to effectively partition and select point clouds remain a challenge. Many approaches~\cite{klokov2017escape,jiang2018pointsift,huang2018recurrent,li2018so} explore new grouping and sampling strategies. In~\cite{liu2018point2sequence,cheraghian20193dcapsule}, new modules are added to original PointNet++ in order to gain a better performance.

Graph Neural Networks (GNNs)~\cite{scarselli2009graph} have been widely used to deal with irregular data structure. There are also a bunch of works trying to apply GNNs to solve point cloud processing problem. Those approaches~\cite{wang2018deep,wang2018dynamic,qi20173d,li2015gated,xu2018spidercnn,shen2018mining,li2018pointcnn} usually build local graphs in the neighborhood of the Euclidean or feature space, utilize MLPs as continuous convolutional kernel functions, and aggregate local features by weighted sum or pooling from neighborhood to center. DGCNN~\cite{wang2018dynamic} proposes an EdgeConv operation which concatenates central and neighboring point features and learns new features by MLP and maxpooling. 3DGNN~\cite{qi20173d} applies gated graph neural networks~\cite{li2015gated} on semantic segmentation task. SpiderCNN~\cite{xu2018spidercnn} defines a continuous kernel function as a product of step function and a Taylor polynomial. KCNet~\cite{shen2018mining} proposes a kernel correlation and graph pooling layer to exploit local structures. PointCNN~\cite{li2018pointcnn} applies $\chi$-transform operator on local graphs.

GNN-based methods still have some problems. First, the graph construction process based on K-nearest neighbors (KNN) is sensitive to point cloud density. Second, using MLPs to directly learn from point coordinates is inefficient, as it ignores some explicitly defined geometric relations. Different from those methods, our approach is not sensitive to point cloud density due to the proposed normalization term, and geometric relations between discrete kernel weights and point clouds are explicitly defined by an interpolation function.

\section{Method}

In this section, we first revisit convolutions on different types of point sets. We then introduce our proposed Interpolated Convolution operation (InterpConv) and key elements of the InterpConv algorithm. Finally, we show details for our network architectures for 3D object recognition and semantic segmentation.

\subsection{Convolutions on Point Sets}
Standard 2D and 3D convolutions have achieved great success in handling regularly-arranged data such as images and voxel grids. When it comes to sparse and irregular point sets such as 3D point clouds, multiple variants of convolutions have been proposed. In this section, we review those convolutions to motivate the design of our InterpConv operation.

Considering a standard 3D convolution, let 3D voxel grids or features be denoted by $F : \mathbb{Z}^{3} \rightarrow \mathbb{R}^{c}$, and the convolutional kernel weights $W$ be a series of $1 \times c$ weight vectors, where $c$ is the number of channels. Standard convolution at location $\hat{p}$ can be formulated as \begin{equation}\label{3.1.1}
F \ast W(\hat{p})= \sum_{p^{\prime}\in\Omega} F(\hat{p}+p^{\prime})\cdot W(p^{\prime}),
\end{equation}where $\Omega = \{p^{\prime}\in\mathbb{Z}^{3}:(-n,-n,-n),\cdots,(n,n,n)\}$ is the set of kernel weight vectors' coordinates relative to the kernel center. The kernel size is assumed to be $2n+1$, and $\cdot$ denotes dot production between two vectors. 


When it comes to irregular inputs, points are no longer regularly-arranged and distances between points become irregular. Some approaches~\cite{wang2018deep,wu2018pointconv} adopt a continuous weight function $W(p_{\delta})$, which takes relative coordinates $p_{\delta}$ of neighboring points $\hat{p}+p_{\delta}$ to the central point $\hat{p}$ as inputs, to predict the convolutional weights. The continuous function $W(p_{\delta})$ is no longer a $1 \times c$ weight vector but a mapping $\mathbb{R}^{3}\rightarrow\mathbb{R}^{c}$ commonly implemented by MLPs. Then the continuous convolution can be formulated as\begin{equation}\label{3.1.2}
F \ast W(\hat{p})= \sum_{p_{\delta}} F(\hat{p}+p_{\delta})\cdot W(p_{\delta}).
\end{equation}It is worth noting that applying graph neural networks~\cite{wang2018dynamic,xu2018spidercnn,shen2018mining} to handle point clouds essentially shares the same idea with continuous convolutions.

Replacing discrete kernel weights $W(p^{\prime})$ with continuous functions $W(p_{\delta})$ remains some problems. Simply learning continuous functions by MLPs cannot always work in practice~\cite{xu2018spidercnn}. The predicted parameters might be too many, and the learning process is inefficient and sometimes unstable. Knowledge on the great success of discrete kernels in images cannot be transfered to point clouds recognition tasks as well. 

\subsection{Interpolated Convolution for 3D Point Clouds}

In our approach, we adopt the design of discrete convolutional weights while maintaining the characteristics of continuous distances, by decoupling $W(p_{\delta})$ into two parts: spatially-discrete kernel weights $W(p^{\prime})\in\mathbb{R}^{c}$ and an interpolation function $T(p_{\delta},p^{\prime})$. We note that a spatially-discrete kernel weight $W(p^{\prime})$ is a $1 \times c$ vector, which can be initialized and updated during training, and $p^{\prime}$ is the relative coordinate of this kernel weight vector to the kernel center. The interpolation function $T(p_{\delta},p^{\prime}) : \mathbb{R}^{3}\times\mathbb{R}^{3} \rightarrow \mathbb{R}$ takes the coordinate of a kernel weight vector $p^{\prime}$ and a neighboring input point $p_{\delta}$ as inputs, and computes a weight by certain interpolation algorithm. Our approach takes every input point in the neighborhood of a kernel weight vector into account. In order to make convolutions sparsity invariant, a density normalization term $N_{p^{\prime}}$, which sums the interpolation weights or number of input points in the neighborhood of $p^{\prime}$, is needed for each kernel weight vector $W(p^{\prime})$. Finally, our InterpConv centered at location $\hat{p}$ can be formulated as\begin{equation}\label{3.1.3}
F\ast W(\hat{p})=\sum_{p^{\prime}}\frac{1}{N_{p^{\prime}}}\sum_{p_{\delta}}T(p_{\delta},p^{\prime})\\
F(\hat{p}+p_{\delta})\cdot W(p^{\prime}).
\end{equation}We note that unlike standard convolutions, where kernel weights are regularly-arranged, kernel-weight coordinates $p^{\prime}$ in InterpConvs can be set flexibly or even learned during training.

There are three key parts of our proposed InterpConv operation: spatially-discrete kernel weights $W$, an interpolation function $T$, and a normalization term $N$. We first discuss those three parts separately, and then introduce the complete algorithm.

\textbf{Discrete kernel weights.} In 2D convolution~\cite{krizhevsky2012imagenet}, one kernel can be represented as an $n \times n \times c$ tensor, where $n$ denotes the kernel size and $c$ denotes the number of channels. In~\cite{chen2018deeplab,dai2017deformable}, one kernel is split into $n \times n$ weight vectors, each of which is of size $1 \times c$. By doing so, kernel weights no longer have to be regularly-arranged, but can be flexibly placed on 2D grids. 

In our approach, we further improve this idea by defining a set of kernel weight vectors for each convolutional kernel in the 3D Euclidean space. Each kernel weight vector $W(p^{\prime})$ has a 3D coordinate $p^{\prime}$ to store its location relative to the kernel center, and its weights are stored in a $1 \times c$ vector, which will be initialized and updated during training. The vector coordinate $p^{\prime}$ can either be fixed or updated during training. To simplify the problem, we fix kernel-weight coordinates in most experiments and organize them as a cube, namely, kernel weight vectors are arranged at $3 \times 3 \times 3$ 3D regular grids if the total number of kernel weight vectors is $27$. We note that this is an analogy of standard $3 \times 3 \times 3$ discrete convolutions while kernel weight vectors can theoretically be placed at arbitrary locations in the 3D space.

As we arrange kernel weight vectors as a cube, we define two important hyperparameters: the kernel size $n \times n \times n$ and kernel length $l$. The coordinate set of spatially-discrete kernel weight vectors can be formulated as\begin{equation}\label{3.2.1}
\begin{split}
\Phi=\Big\{(x^{\prime},y^{\prime},z^{\prime}) \Big| & x^{\prime},y^{\prime},z^{\prime}=kl,\\& 
k\in\Big\{-\frac{n-1}{2},\cdots,\frac{n-1}{2}\Big\}\Big\},
\end{split}
\end{equation} where $p^{\prime}=(x^{\prime},y^{\prime},z^{\prime})$. Similar to the definition of kernel size in standard convolutions, kernel size $n \times n \times n\in\mathbb{Z}^{3}$ means that there are $n$ kernel weight vectors on each edge of one kernel, and the total number of kernel weight vectors is $n^{3}$. Kernel length $l\in\mathbb{R}$ is the distance between two adjacent weight vectors. It determines the actual 3D size of a kernel in the Euclidean space and is defined to control the receptive field, from which one convolutional kernel can ``see'' input point clouds. If $l$ is small, the convolutional kernel is able to capture fine-grained local structures, otherwise it encodes more global shape information. 

\textbf{Interpolation functions.} One problem to apply discrete kernels on irregular point clouds is that kernel weight vectors' spatial locations generally do not align with input points. Naively rasterizing point clouds into regular grids~\cite{maturana2015voxnet,hua2018pointwise} solves part of the problem, but at the cost of losing local structures. In our approach, we solve this problem while keep all fine-grained structures by adopting an interpolation function. Namely, we first find a set of input points near each kernel weight vector, and then interpolate their features to be assigned to the kernel weight vectors for convolution. We propose two interpolation functions: trilinear interpolation and Gaussian interpolation.

Trilinear interpolation is a commonly-used method to approximate the value of an intermediate point in a 3D grid by values of adjacent lattice points. The intermediate point's value is calculated by a weighted sum of lattice points' values, and the weights characterize closeness between intermediate and lattice points. In our approach, we adopt the inverse process of trilinear interpolation. Namely, we first compute the weights that lattice points (kernel-weight coordinates) contribute to the intermediate point (input point) and then we inversely assign the input point feature to adjacent kernel-weight coordinates with those weights.

For trilinear interpolation, we find $8$ adjacent kernel-weight coordinates $p^{\prime}$ for each input point $p_{\delta}$ in the kernel, and then we normalize input point and kernel weights into a unit-length cube. Finally we compute the trilinear interpolation weights by\begin{equation}\label{3.2.2}
T_{tr}(p_{\delta}, p^{\prime})=(1-|x_{\delta}-x^{\prime}|)(1-|y_{\delta}-y^{\prime}|)(1-|z_{\delta}-z^{\prime}|),
\end{equation}
where input point $p_{\delta}=(x_{\delta},y_{\delta},z_{\delta})$ is the relative point coordinate to the kernel center, and kernel-weight coordinate $p^{\prime}=(x^{\prime},y^{\prime},z^{\prime})$. We further note that Eq.~(\ref{3.2.2}) is a simplified format for normalized points. One property of trilinear interpolation is self-normalization, namely, all $8$ weights which an input point assigns to can sum up to $1$. 

In Gaussian interpolation, we assign each input point $p_{\delta}$ to each kernel weight vector at $p^{\prime}$ with a weight factor calculated by the following Gaussian function,\begin{equation}\label{3.2.3}
T_{G}(p_{\delta}, p^{\prime})=e^{-\frac{(x_{\delta}-x^{\prime})^{2}+(y_{\delta}-y^{\prime})^{2}+(z_{\delta}-z^{\prime})^{2}}{2\sigma^{2}}},
\end{equation}where the hyperparameter $\sigma$ controls the decay rate. To save computation, if a 3D point is $3\sigma$ away from a weight vector, its assignment coefficient to the vector is directly set to $0$ and will not be calculated. It is worth noting that other functions, for example, linear basis functions, can also be adopted as the interpolation function. 


\textbf{Normalization terms.} Given the fact that we take all neighboring points of a kernel weight vector into calculation, normalization is necessary to keep convolutions invariant to points density. There are two ways of normalization. We can aggregate and normalize the point features by 
\begin{equation}\label{3.2.4}
f_{aggregate} = \frac{\sum^{N}_{i=1} t_{i}f_{i}}{N},
\end{equation}
where $N$ is the number of neighboring points, $f_{i}$ is the $i$th point feature, and $t_{i}$ denotes its interpolation weight. Apart from normalizing according to the number of points, we can also normalize the sum of interpolation weights:
\begin{equation}\label{3.2.5}
f_{aggregate} = \frac{\sum^{N}_{i=1} t_{i}f_{i}}{\sum^{N}_{i=1} t_{i}}.
\end{equation}

We can perform normalization either on each kernel weight vector or on the whole convolutional kernel. We argue that normalization on each kernel weight vector is more accurate, since input points are not uniformly distributed in the whole kernel. 

\begin{algorithm}[t]\label{a1}
	\caption{The InterpConv Algorithm}
	\hspace*{0.02in} {\bf Input:} 
	point coordinates $\mathit{p}\in\mathbb{R}^{3}$, point features $\mathit{f}\in\mathbb{R}^{c}$
	\hspace*{0.02in} {\bf Output:} output coordinates $\mathit{\hat{p}}\in\mathbb{R}^{3}$, new features $\mathit{\hat{f}}\in\mathbb{R}^{c^{\prime}}$
	\hspace*{0.02in} {\bf Parameter:} $\mathit{c^{\prime}}$ kernels with $n$ weight vectors $\mathit{w}\in\mathbb{R}^{c}$ and\\ 
	\hspace*{0.02in} shared weight coordinates $\mathit{p^{\prime}}\in\mathbb{R}^{3}$ in each kernel
	\begin{algorithmic}[1]
		\State Sample $\mathit{\hat{p}}$ from $\mathit{p}$ or $\mathit{\hat{p}}\leftarrow\mathit{p}$
		\For{\textbf{each} $\hat{p}$}
		\For{\textbf{each} $p^{\prime}$}
		\For{\textbf{each} neighboring $p$, feature $f_{p}$}
		\State$\mathit{p}_{\delta}\leftarrow p - \hat{p}$ 
		\State
		$\mathit{t}\leftarrow\mathit{T}({p}^{\prime},p_{\delta})$
		\State $f_{i}\leftarrow\mathit{f_{i}}+\mathit{t}\mathit{f_{p}}$
		\EndFor
		\State $\mathit{f_{i}}\leftarrow Normalize( \mathit{f_{i}} )$
		\EndFor
		\State
		$F\leftarrow[f_{1},\cdots,f_{n}]$
		
		\For{\textbf{each} kernel $k$}
		\State
		$W_{k}\leftarrow[w_{1}^{k},\cdots,w_{n}^{k}]$
		\State
		$v_{k}\leftarrow F\cdot W_{k}$		
		\EndFor
		\State
		$\hat{f}\leftarrow[v_{0},\cdots,v_{c^{\prime}}] $
		\EndFor
		
		\State 
		\Return{$\hat{p},\hat{f}$}
	\end{algorithmic}
\end{algorithm}

\textbf{The InterpConv Algorithm.} An InterpConv operation takes point cloud coordinates and their features as input, and outputs new point coordinates and features. We note that output point coordinates can be set as the same as the input points, or downsampled from input point clouds. The center of convolutional kernels is placed at each output point coordinate, and kernel-weight coordinates are further determined by the relative coordinates of the weight vectors following Eq.~(\ref{3.2.1}). We calculate interpolation weights between kernel weight vectors and adjacent input points, and then aggregate feature by the weighted sum of all neighboring point features. The aggregated features are further normalized to keep it sparsity invariant. Finally dot production is applied between the normalized features and kernel weight vectors. A convolutional kernel sums all the results and $c^{\prime}$ kernels constitute a $1 \times c^{\prime}$ new feature vector at the output coordinate. See the InterpConv algorithm for details.

\begin{figure*} 
	\centering 
	\subfigure[Classification network]{
		\includegraphics[width=0.45\linewidth]{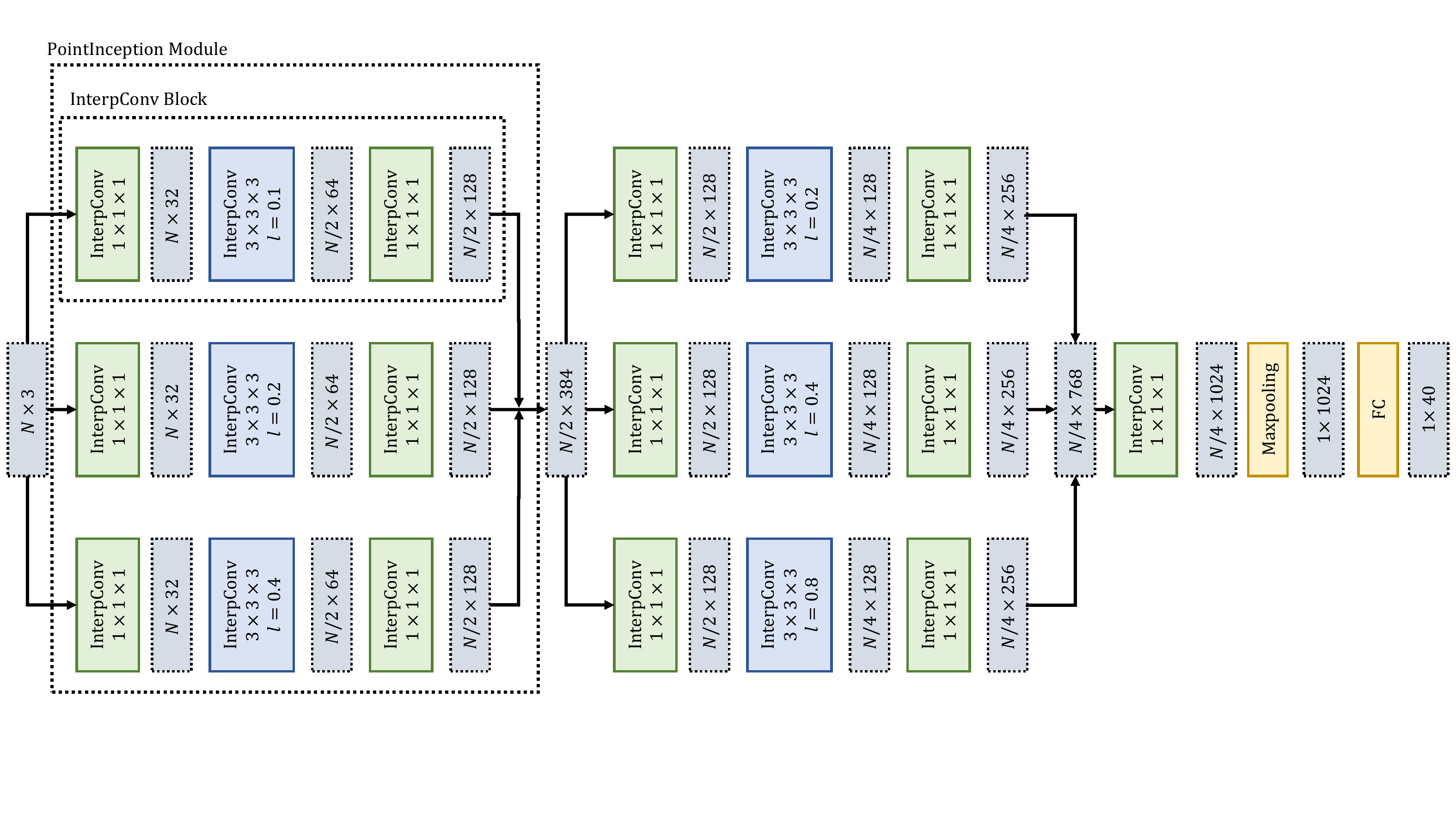}\label{fig3.1}}
	\hspace{0.01\linewidth}
	\subfigure[Segmentation network]{	
		\includegraphics[width=0.45\linewidth]{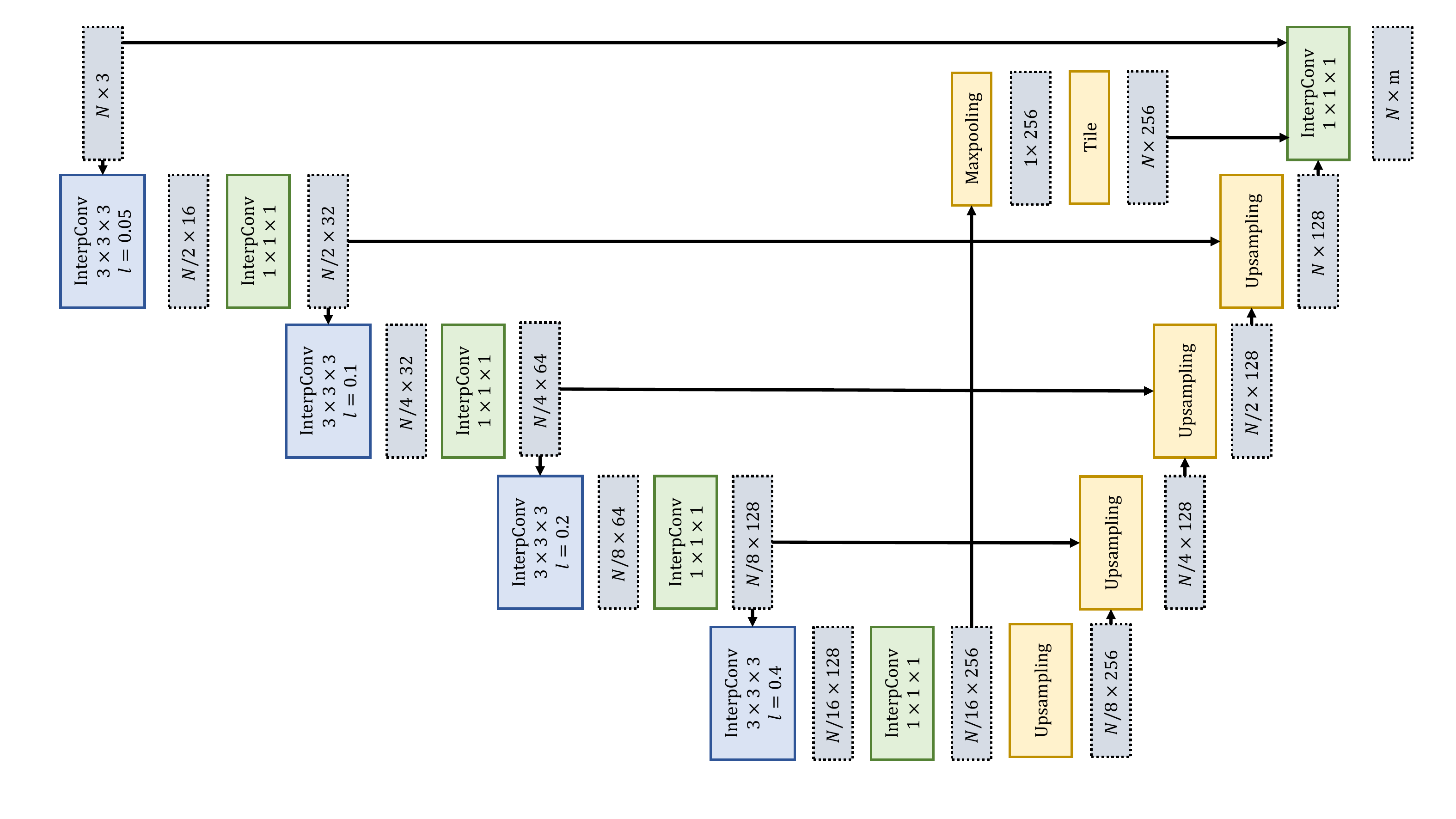}\label{fig3.2}}	
	\caption{Interpolated Convolutional Neural Networks 
		(InterpCNNs). Gray boxes indicate the size of input and output data, and other boxes are all network layers. In the classification network \subref{fig3.1}, we extend the idea of Inception module~\cite{szegedy2015going} and stack two multi-branch, multi-receptive-field PointInception modules to capture both local and context geometric information. We note that kernel length $l$ varies at different branches. In the segmentation network \subref{fig3.2}, we share similar spirit as U-Net~\cite{ronneberger2015u} and build an InterpConv-based deep encoder-decoder architecture. Kernel length $l$ begins with a small value and becomes larger as the network goes deeper.}	
	\label{fig3}
\end{figure*}

\subsection{Network Architectures}
In this section, we introduce details for two deep architectures based on our InterpConv approaches. We explore embedding mutli-scale context features in the classification network and a deep encoder-decoder architecture in the segmentation network. See Figure \ref{fig3} for details.

The classification network consists of a series of InterpConv blocks which is mainly composed of three InterpConv layers. In the InterpConv block, the first and last layer are of kernel size $1 \times 1 \times 1$ and the middle layer has a kernel size $3 \times 3 \times 3$. The first InterpConv layer reduces channel dimensions and the last InterpConv layer increases channel dimensions, leaving the middle InterpConv layer with relatively small input and output channels. One BatchNorm~\cite{ioffe2015batch} and ReLU~\cite{xu2015empirical} layer also follow each InterpConv layer in the block.  

Apart from this, we propose the PointInception module to encode multi-scale geometric features. Similar to the Inception module~\cite{szegedy2015going} in 2D CNNs, our PointInception module also concatenates multi-branch features. However, we design each branch as one InterpConv block with a different kernel length $l$. The hyperparameter $l$ determines the distances between adjacent kernel weight vectors in the Euclidean space and controls the receptive field of the convolution. So the PointInception module is able to capture both fine-grained local structures and shape context information by combining multi-branch outputs. We further explore a deeper model by stacking two PointInception modules.

In the segmentation network, we share the similar spirit as U-Net~\cite{ronneberger2015u} and build a deep encoder-decoder architecture. We stack multiple $3 \times 3 \times 3$ InterpConv layers in the encoder, and in each layer output points are downsampled. In the first $3 \times 3 \times 3$ InterpConv layer, we set the kernel length $l$ as a small value in order to capture fine-grained geometric structures, which is important in semantic segmentation. We then gradually enlarge the kernel length $\mathit{l}$ in the following blocks to capture context information. For the upsampling layers in the decoder, we utilize feature propagation layers following~\cite{qi2017pointnet++}. Skip connections are added between layers that have the same number of output points. The decoder outputs are then fed into an InterpConv layer with kernel size $1 \times 1 \times 1$ to obtain the final predictions.
\section{Experiments}
In this section, we evaluate the efficacy of Interpolated Convolutional Neural Networks on multiple tasks, including shape classification, object part segmentation and indoor scene semantic parsing. In all experiments, we implement the models using CUDA and PyTorch on NVIDIA TITAN X GPUs, and we use the Adam optimizer. We first demonstrate the performances of our approach on those tasks. Then we discuss key components of our method in ablation study.

\subsection{Shape Classification}

\textbf{Dataset.} We evaluate the 3D shape classification performance of our network on the benchmark dataset ModelNet40~\cite{chang2015shapenet}. ModelNet40 is composed of 12,311 CAD models which belong to 40 categories with 9,843 for training and 2,468 for testing. We use the point cloud conversion of ModelNet40, where 2,048 points are sampled from each CAD model. We further sample 1,024 points for training and testing following~\cite{qi2017pointnet}.


\begin{table}
	
	\begin{center}
		\begin{tabular*}{\linewidth}{@{\extracolsep{\fill}}c c c}
			\hline
			& Input & Acc. \\
			\hline
			Subvolume~\cite{qi2016volumetric} & voxels & 89.2\%\\
			VRN Single~\cite{brock2016generative} & voxels & 91.3\%\\
			OctNet~\cite{riegler2017octnet} & hybrid grid octree & 86.5\%\\
			ECC~\cite{simonovsky2017dynamic} & graphs & 87.4\%\\
			\hline
			PointwiseCNN~\cite{hua2018pointwise} & 1024 points & 86.1\%\\
			PointNet~\cite{qi2017pointnet} & 1024 points & 89.2\%\\
			PointNet++~\cite{qi2017pointnet++} & 1024 points & 90.7\%\\
			PointNet++~\cite{qi2017pointnet++} & 5000 points+normal & 91.9\%\\
			Kd-Network~\cite{klokov2017escape} & 1024 points & 91.8\%\\
			ShapeContextNet~\cite{xie2018attentional} & 1024 points & 90.0\%\\
			KCNet~\cite{shen2018mining} & 1024 points & 91.0\%\\
			PointCNN~\cite{li2018pointcnn} & 1024 points & 92.2\%\\
			DGCNN~\cite{wang2018dynamic} & 1024 points & 92.2\%\\
			SO-Net~\cite{li2018so} & 2048 points & 90.9\%\\
			SpiderCNN~\cite{xu2018spidercnn} & 1024 points+normal & 92.4\%\\
			Point2Seq~\cite{liu2018point2sequence} & 1024 points & 92.6\%\\
			3DCapsule~\cite{cheraghian20193dcapsule} & 1024 points & 92.7\%\\
			PointConv~\cite{wu2018pointconv} & 1024 points+normal & 92.5\%\\
			\hline
			InterpCNN (ours) & 1024 points & \textbf{93.0\%}\\
			\hline
		\end{tabular*}
	\end{center}
	\caption{Classification results on ModelNet40. Overall accuracy is reported.}
	\label{t1}
\end{table}

\textbf{Implementation details.} We adopt the classification network in Figure~\ref{fig3.1}. We use Gaussian interpolation as our interpolation function and fix the Gaussian bandwidth $3\sigma$ to 0.1 in all InterpConv blocks. Point clouds are downsampled to half of the input number after each $3 \times 3 \times 3$ InterpConv layer. The input point clouds are randomly scaled by a factor ranging from 0.8 to 1.2, then jittered by a zero-mean Gaussian noise with 0.02 standard deviation. We trained the network for 480 epoches with initial learning rate 0.001 and decay rate 0.7 every 80 epoches with batch size $16$.

\textbf{Results.} We report the overall accuracy on this dataset. In Table~\ref{t1}, we compare our InterpCNN with other approaches. We demonstrate that the deep architecture based on InterpConvs performs much better than graph-based and voxel-based counterparts, with 0.8\% improvement on the best graph-based network DGCNN~\cite{wang2018dynamic}. Our approach performs even better than Point2Seq~\cite{liu2018point2sequence} and 3DCapsule~\cite{cheraghian20193dcapsule} in which many modules and model compacity are added on top of PointNet++~\cite{qi2017pointnet++} to gain a better performance.

\subsection{Object Part Segmentation}

\textbf{Dataset.} We evaluate our segmentation network on the part segmentation dataset ShapeNet Parts~\cite{yi2016scalable}. ShapeNet Parts consists of 16,880 models from 16 shape categories, with 14,006 for training and 2,874 for testing. Each model is annotated with 2 to 6 parts and there are 50 different parts in total. Each point sampled from the models is annotated with a part label.

\begin{table}
	
	\begin{center}
		\begin{tabular*}{\linewidth}{@{\extracolsep{\fill}}c c c}
			\hline
			& Cat. & Ins. \\
			& mIOU & mIOU \\
			\hline
			PointNet~\cite{qi2017pointnet} & 80.4\% & 83.7\%\\
			PointNet++~\cite{qi2017pointnet++} & 81.9\% & 85.1\%\\
			FCPN~\cite{rethage2018fully} & - & 84.0\%\\
			SyncSpecCNN~\cite{yi2017syncspeccnn} & 82.0\% & 84.7\%\\
			SSCN~\cite{graham20183d} & 83.3\% & 86.0\%\\
			SPLATNet~\cite{su2018splatnet} & 83.7\% & 85.4\%\\
			SpiderCNN~\cite{xu2018spidercnn} & 81.7\% & 85.3\%\\
			SO-Net~\cite{li2018so} & 81.0\% & 84.9\%\\
			PCNN~\cite{atzmon2018point} & 81.8\% & 85.1\%\\
			KCNet~\cite{shen2018mining} & 82.2\% & 83.7\%\\
			ShapeContextNet~\cite{xie2018attentional} & - & 84.6\%\\
			SpecGCN~\cite{wang2018local} & - & 85.4\%\\
			3DmFV~\cite{ben20173d} & 81.0\% & 84.3\%\\
			RSNet~\cite{huang2018recurrent} & 81.4\% & 84.9\%\\
			PointCNN~\cite{li2018pointcnn} & 84.6\% & 86.1\%\\
			DGCNN~\cite{wang2018dynamic} & 82.3\% & 85.1\%\\
			SGPN~\cite{wang2018sgpn} & 82.8\% & 85.8\%\\
			PointConv~\cite{wu2018pointconv} & 82.8\% & 85.7\%\\
			Point2Seq~\cite{liu2018point2sequence} & - & 85.2\%\\
			\hline
			InterpCNN (ours) & 84.0\% & \textbf{86.3\%}\\
			\hline
		\end{tabular*}
	\end{center}
	\caption{Segmentation results on ShapeNet Parts. Mean IoU over categories (Cat.) and instances (Ins.) are reported.}
	\label{t2}
\end{table}

\textbf{Implementation details.} We use the segmentation network in Figure~\ref{fig3.2}. During training we randomly sample 2,048 points from each object and use the original point clouds for testing. Different from the classification network, we utilize a trilinear interpolation function with a smaller kernel length $l$, which is shown to perform much better. The kernel length $l$ starts with 0.05 in the first InterpConv layer and doubles in the following layers. We use a minibatch of 32 in each GPU and 4 GPUs to train a model. We set the initial learning rate to be 0.005. Data augmentation is the same as classification.

\textbf{Results.} We report mean IOU over categories and instances in Table~\ref{t2}. It is worth noting that mean IOU over instances is more realistic. Our approach performs better than compared methods on mean IOU over instances.
\begin{figure} 
	\centering 
	\includegraphics[width=1\linewidth]{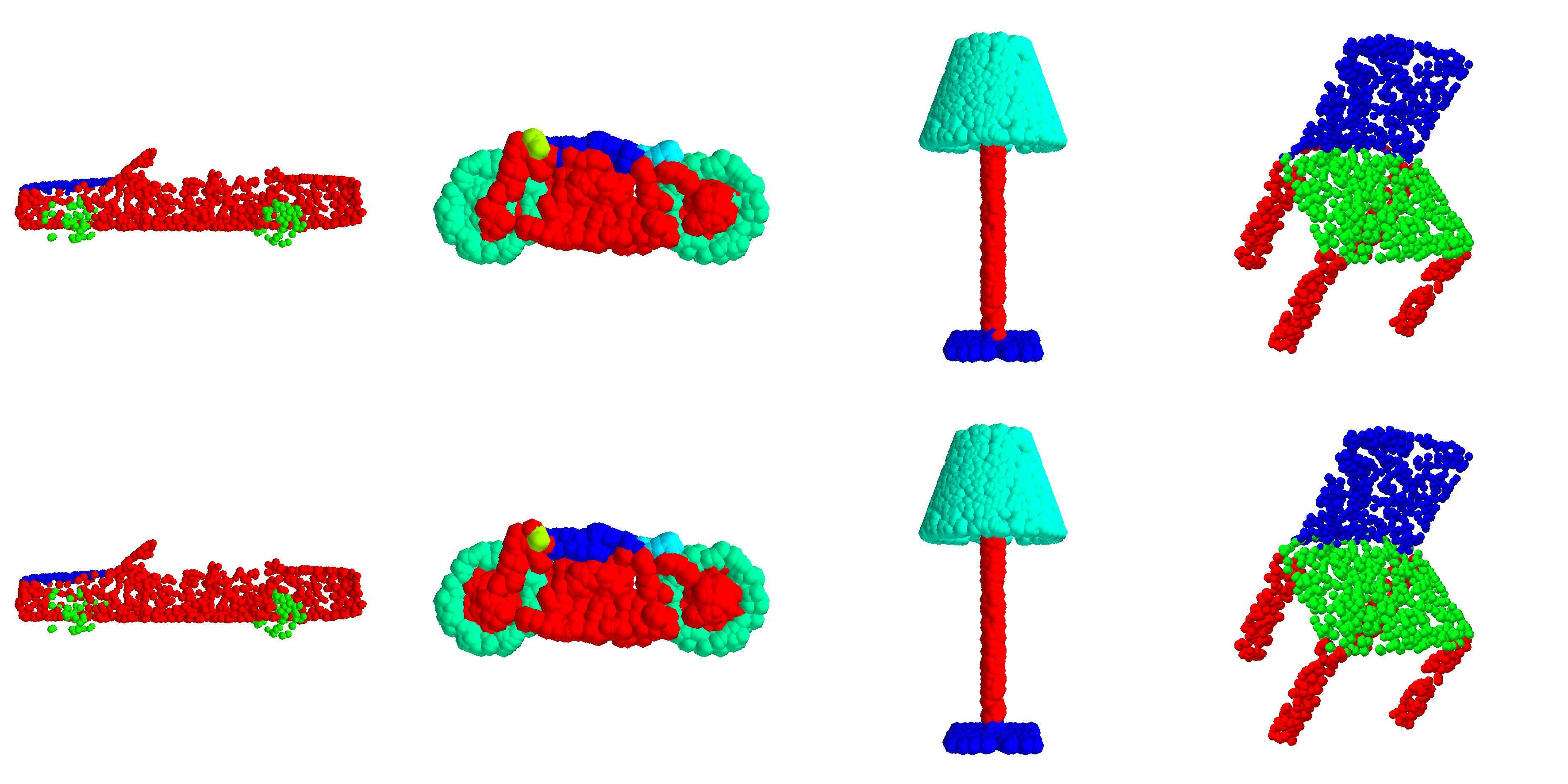}		
	\caption{Visualization of object part segmentation results on ShapeNet Parts. The first row is ground truth and the second row is our predictions. From left to right are cars, motorbikes, lamps and chairs.}	
	\label{fig6}
\end{figure}


\subsection{Indoor Scene Segmentation}

\textbf{Dataset.} S3DIS~\cite{armeni20163d} is an indoor sence semantic parsing dataset which contains 271 rooms in 6 areas. Each room is scanned by Matterport scanners and every point in the scan is annotated with one semantic label from 13 categories. We follow~\cite{qi2017pointnet} and split rooms into $1$m $\times$ $1$m blocks for training and testing.

\textbf{Implementation details.} Similar to the part segmentation task we use the same architecture in Figure~\ref{fig3.2}. The difference is that we take 4,096 points from each $1$m $\times$ $1$m block as inputs during training. We construct a 9D vector (XYZ, RGB, and the normalized location) for each input. Other configurations are the same as that in the object part segmentation task.  

\begin{table}
	
	\begin{center}
		\begin{tabular*}{\linewidth}{@{\extracolsep{\fill}}c c c}
			\hline
			& Overall & Cat. \\
			& Accuracy & mIOU \\
			\hline
			PointNet~\cite{qi2017pointnet} & 78.5\% & 47.6\%\\
			ShapeContextNet~\cite{xie2018attentional} & 81.6\% & 52.7\%\\
			RSNet~\cite{huang2018recurrent} & - & 56.5\%\\
			PointCNN~\cite{li2018pointcnn} & 88.1\% & 65.4\%\\			
			DGCNN~\cite{wang2018dynamic} & 84.1\% & 56.1\%\\		
			SGPN~\cite{wang2018sgpn} & 80.8\% & 50.4\%\\
			SPGraph~\cite{landrieu2018large} & 85.5\% & 62.1\%\\
			\hline
			InterpCNN (ours) & \textbf{88.7\%} & \textbf{66.7\%}\\
			\hline
		\end{tabular*}
	\end{center}
	\caption{6-fold validation results on S3DIS. Overall accuracy and mean IOU over categories are reported.}
	\label{t3}
\end{table}

\textbf{Results.} Following~\cite{qi2017pointnet}, we adpot 6-fold validations on 6 areas, and we report overall accuracy and mean IOU over categories in Table~\ref{t3}. Our approach significantly outperforms state-of-the-art methods on both accuracy and mean IOU.

\begin{figure} 
	\centering 
	\includegraphics[width=1\linewidth]{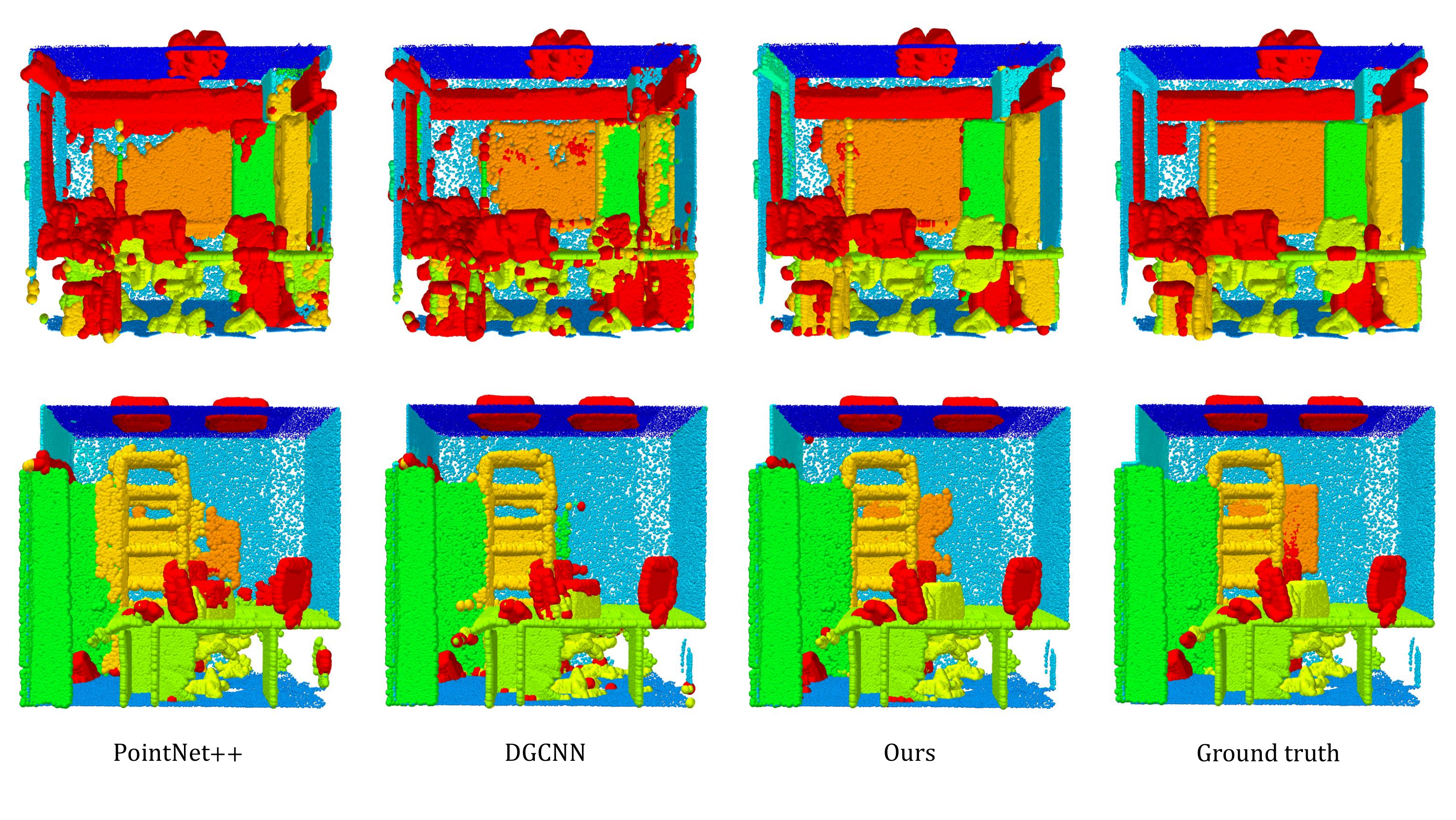}		
	\caption{Qualitative evaluation on S3DIS compared with PointNet++ and DGCNN.}	
	\label{fig7}
\end{figure}

\subsection{Ablation Study}


We perform ablation studies to investigate components of our InterpCNNs on ModelNet40 and ShapeNet Parts.

\textbf{Effectiveness of kernel size $n$ and kernel length $l$.} We explore different settings of hyperparameter $n$ and $l$ for the 1st and 2nd PointInception module in Table~\ref{t4} and Table~\ref{t5}. We first try the setting of all InterpConvs with kernel size $1 \times 1 \times 1$ and only use a maxpooling layer to aggregate global features. We note that this architecture is similar to PointNet~\cite{qi2017pointnet} in which the network cannot capture local structures, and the result is much worse. This indicates the great power of InterpConvs with kernel size more than $1$. Simply replacing one $1 \times 1 \times 1$ InterpConv with $3 \times 3 \times 3$ for each InterpConv block obtains a performance gain of 3\%. We also try InterpConvs with a larger kernel size $5 \times 5 \times 5$ but the performance does not improve. We demonstrate that utilizing $3 \times 3 \times 3$ InterpConvs is sufficiently effective and this also reduces model parameters compared with the $5 \times 5 \times 5$ counterparts. We also explore different kernel lengths and we show that this hyperparameter has a significant effect on the final performance. Either too small or too large the kernel length $l$ will harm the accuracy. 


\textbf{Effectiveness of interpolation functions.} We try both Gaussian and trilinear interpolation functions in all tasks. In Table~\ref{t6}, the results show that Gaussian interpolation performs better in classification while trilinear interpolation is better in segmentaion. We argue that trilinear interpolation can capture fine-grained geometric structures better than Gaussian counterpart, which is more important for segmentation. Gaussian interpolation is able to obtain global shape information more effectively.

\textbf{Effectiveness of normalization methods.} We try normalization according to the number of neightboring points (Eq.~(\ref{3.2.4})) and the sum of interoplation weights (Eq.~(\ref{3.2.5})). The results in Table~\ref{t9} indicate that both two methods are effective and show comparable performances. It is worth noting that in extreme cases where there are only a few close points but many far away points in the neighborhood of a kernel-weight coordinate, normalization on the sum of interpolation weights is more appropriate.  
\begin{table}
	\begin{center}
		\begin{tabular*}{\linewidth}{@{\extracolsep{\fill}}c c c}
			\hline
			1st module & 2nd module & Accuracy \\
			\hline
			$1 \times 1 \times 1$ & $1 \times 1 \times 1$ & 89.9\%\\
			$3 \times 3 \times 3$ & $5 \times 5 \times 5$ & 92.9\%\\
			$5 \times 5 \times 5$ & $3 \times 3 \times 3$ & 92.8\%\\
			$3 \times 3 \times 3$ & $3 \times 3 \times 3$ & \textbf{93.0\%} \\
			\hline
		\end{tabular*}
	\end{center}
	\caption{Results of different kernel sizes on ModelNet40.}
	\label{t4}
\end{table}

\begin{table}
	\begin{center}
		\begin{tabular*}{\linewidth}{@{\extracolsep{\fill}}c c c}
			\hline
			1st module & 2nd module & Accuracy \\
			\hline
			$0.05-0.1-0.2$ & $0.1-0.2-0.4$ & 92.4\%\\
			$0.1-0.2-0.4$ & $0.2-0.4-0.8$ & \textbf{93.0\%} \\
			$0.2-0.4-0.8$ & $0.4-0.8-1.6$ & 91.9\%\\
			\hline
		\end{tabular*}
	\end{center}
	\caption{Results of different kernel lengths on ModelNet40.}
	\label{t5}
\end{table}

\begin{table}
	\begin{center}
		\begin{tabular*}{\linewidth}{@{\extracolsep{\fill}}c c c}
			\hline
			Interpolation function & ModelNet40 & ShapeNet Parts\\
			\hline
			Gaussian & \textbf{93.0\%} & 85.0\%\\
			Trilinear & 92.5\% & \textbf{86.3\%}\\
			\hline
		\end{tabular*}
	\end{center}
	\caption{Results of different interpolation functions on ModelNet40 and ShapeNet Parts.}
	\label{t6}
\end{table}

\begin{table}
	\begin{center}
		\begin{tabular*}{\linewidth}{@{\extracolsep{\fill}}c c c c}
			\hline
			&Normalization method & Accuracy&\\
			\hline
			&interpolation weights & 92.8\%& \\
			&number of points & \textbf{93.0}\%& \\
			\hline
		\end{tabular*}
	\end{center}
	\caption{Results of different normalization methods on ModelNet40.}
	\label{t9}
\end{table}

\begin{table}
	\begin{center}
		\begin{tabular*}{\linewidth}{@{\extracolsep{\fill}}c c c}
			\hline
			Method & Parameters & Accuracy\\
			\hline
			Subvolume~\cite{qi2016volumetric} & 16.6M & 89.2\%\\
			PointNet~\cite{qi2017pointnet} & 3.5M & 89.2\%\\
			PointNet++ (MSG)~\cite{qi2017pointnet++} & 12M & 90.7\%\\
			\hline
			InterpCNN (ours) & 12.8M& \textbf{93.0\%}\\
			\hline
		\end{tabular*}
	\end{center}
	\caption{Model parameters and performance comparisons on ModelNet40.}
	\label{t7}
	
\end{table}
\begin{table}
	\begin{center}
		\begin{tabular*}{\linewidth}{@{\extracolsep{\fill}}c c c}
			\hline
			Method & Inference time & Accuracy\\
			\hline
			PointNet++ (MSG)~\cite{qi2017pointnet++} & 26.8ms & 90.7\%\\
			DGCNN~\cite{wang2018dynamic} & 89.7ms & 92.2\%\\
			\hline
			InterpCNN (ours) & 31.4ms & \textbf{93.0\%}\\
			\hline
		\end{tabular*}
	\end{center}
	\caption{Inference time comparisons on ModelNet40.}
	\label{t8}
\end{table}

\textbf{Model parameters analysis.} We report the number of parameters in our classification network on ModelNet40. Results in Table~\ref{t7} show that even with the comparable model parameters, PointNet++ still performs much worse than our approach. InterpCNNs also have fewer parameters than other 3D convolution methods.

\textbf{Runtime analysis.} We summarize average inference time based on the classification network with batch size $16$, $1024$ points on an NVIDIA TITAN X GPU, and compare it with pioneering work PointNet++ and DGCNN under the same settings. In Table~\ref{t8}, average inference time of our approach is slightly slower than PointNet++, but much faster than graph-based approach DGCNN.

\begin{figure} 
	\centering 
	\includegraphics[width=1\linewidth]{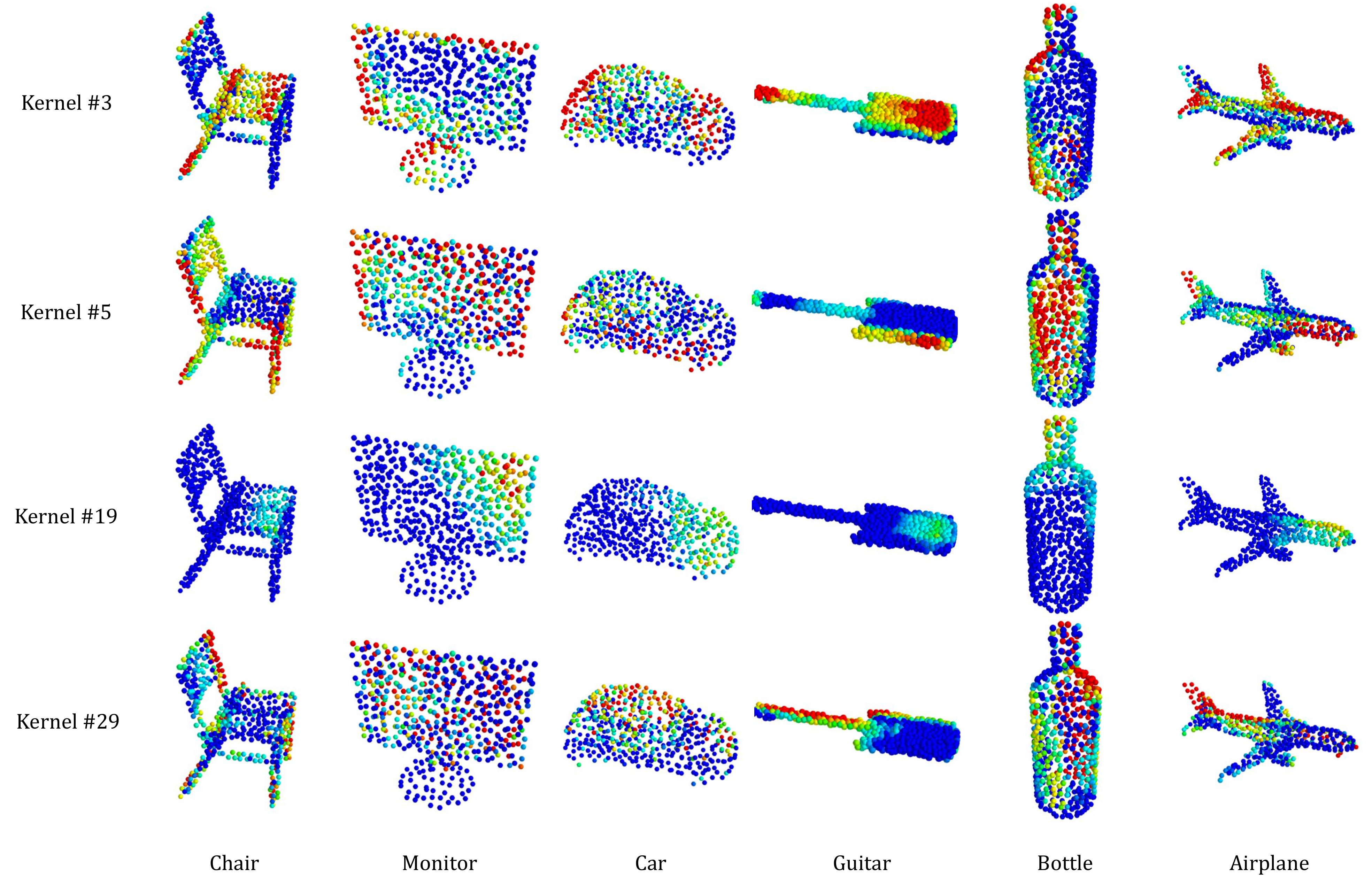}		
	\caption{Visualization of feature activations learned by different InterpConv kernels on ModelNet40.}	
	\label{fig8}
\end{figure}

\begin{figure} 
	\centering 
	\includegraphics[width=1\linewidth]{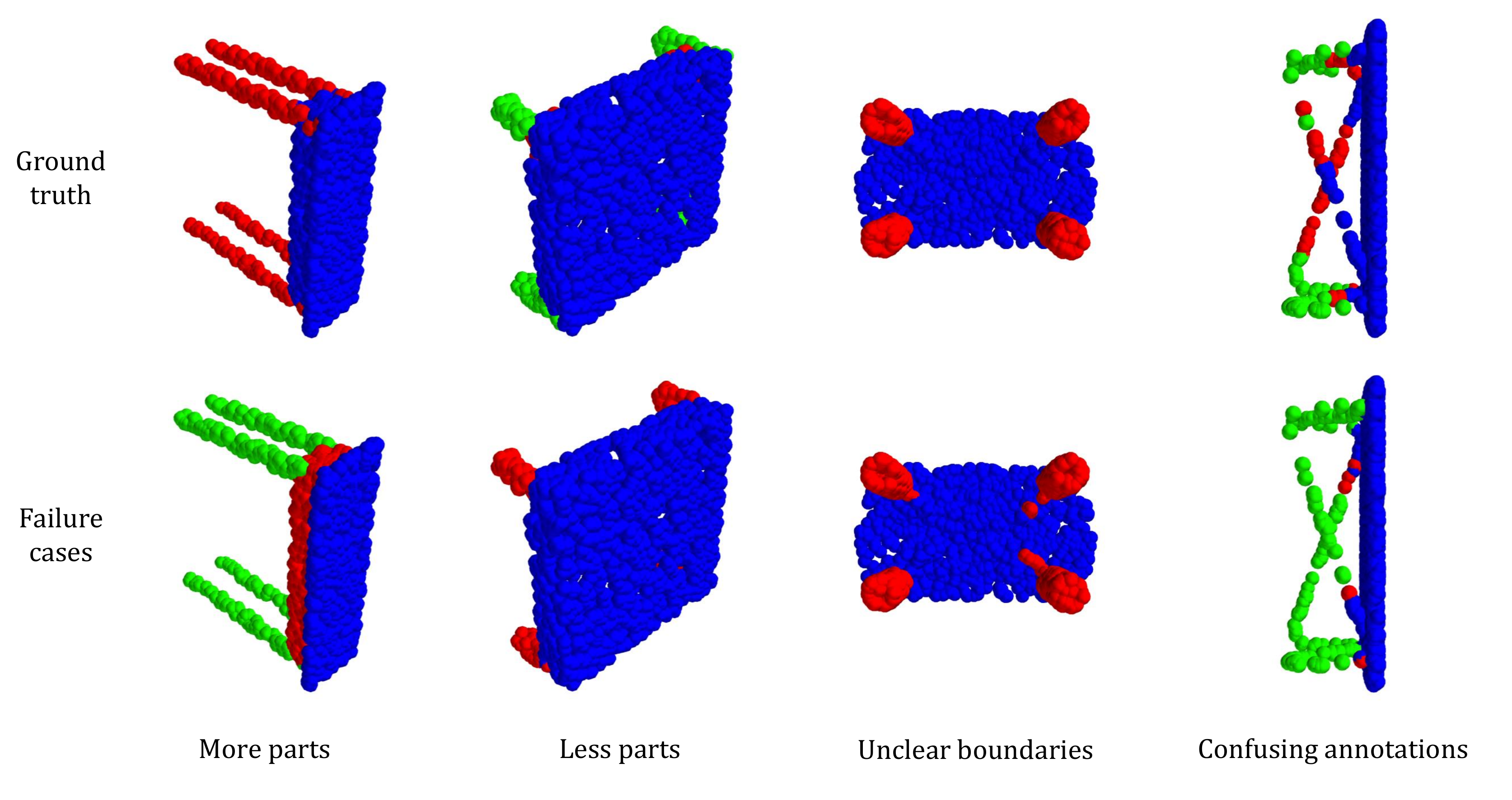}		
	\caption{Visualization of failure cases on ShapeNet Parts.}	
	\label{fig9}
\end{figure}

\textbf{Visualization.} We visualize activations by different kernels in the first $3\times3\times3$ InterpConv layer in Figure~\ref{fig8} and some failure cases in Figure~\ref{fig9}.


\section{Conclusion} 
We propose a novel convolution InterpConv and Interpolated Convolutional Neural Networks (InterpCNNs) for 3D classification and segmentation. Experiments on ModelNet40, ShapeNet Parts and S3DIS show promising results compared with  existing methods. For future work, we plan to explore new deep architectures based on learnable kernel-weight coordinates, and apply our approach on other point cloud processing tasks including 3D detection and instance segmentation.

\section*{Acknowledgements}
This work is supported in part by SenseTime Group Limited, in part by the General Research Fund through the Research Grants Council of Hong Kong under Grants CUHK14202217, CUHK14203118, CUHK14205615, CUHK14207814, CUHK14213616, CUHK14208417, CUHK14239816, in part by CUHK Direct Grant.

{\small
\bibliographystyle{ieee_fullname}
\bibliography{egbib}

\begin{thebibliography}{10}\itemsep=-1pt

\bibitem{armeni20163d}
Iro Armeni, Ozan Sener, Amir~R Zamir, Helen Jiang, Ioannis Brilakis, Martin
  Fischer, and Silvio Savarese.
\newblock 3d semantic parsing of large-scale indoor spaces.
\newblock In {\em Proceedings of the IEEE Conference on Computer Vision and
  Pattern Recognition}, pages 1534--1543, 2016.

\bibitem{atzmon2018point}
Matan Atzmon, Haggai Maron, and Yaron Lipman.
\newblock Point convolutional neural networks by extension operators.
\newblock {\em arXiv preprint arXiv:1803.10091}, 2018.

\bibitem{ben20173d}
Yizhak Ben-Shabat, Michael Lindenbaum, and Anath Fischer.
\newblock 3d point cloud classification and segmentation using 3d modified
  fisher vector representation for convolutional neural networks.
\newblock {\em arXiv preprint arXiv:1711.08241}, 2017.

\bibitem{brock2016generative}
Andrew Brock, Theodore Lim, James~M Ritchie, and Nick Weston.
\newblock Generative and discriminative voxel modeling with convolutional
  neural networks.
\newblock {\em arXiv preprint arXiv:1608.04236}, 2016.

\bibitem{chang2015shapenet}
Angel~X Chang, Thomas Funkhouser, Leonidas Guibas, Pat Hanrahan, Qixing Huang,
  Zimo Li, Silvio Savarese, Manolis Savva, Shuran Song, Hao Su, et~al.
\newblock Shapenet: An information-rich 3d model repository.
\newblock {\em arXiv preprint arXiv:1512.03012}, 2015.

\bibitem{chen2018deeplab}
Liang-Chieh Chen, George Papandreou, Iasonas Kokkinos, Kevin Murphy, and Alan~L
  Yuille.
\newblock Deeplab: Semantic image segmentation with deep convolutional nets,
  atrous convolution, and fully connected crfs.
\newblock {\em IEEE transactions on pattern analysis and machine intelligence},
  40(4):834--848, 2018.

\bibitem{chen2017multi}
Xiaozhi Chen, Huimin Ma, Ji Wan, Bo Li, and Tian Xia.
\newblock Multi-view 3d object detection network for autonomous driving.
\newblock In {\em Proceedings of the IEEE Conference on Computer Vision and
  Pattern Recognition}, pages 1907--1915, 2017.

\bibitem{cheraghian20193dcapsule}
Ali Cheraghian and Lars Petersson.
\newblock 3dcapsule: Extending the capsule architecture to classify 3d point
  clouds.
\newblock In {\em 2019 IEEE Winter Conference on Applications of Computer
  Vision (WACV)}, pages 1194--1202. IEEE, 2019.

\bibitem{dai2017deformable}
Jifeng Dai, Haozhi Qi, Yuwen Xiong, Yi Li, Guodong Zhang, Han Hu, and Yichen
  Wei.
\newblock Deformable convolutional networks.
\newblock In {\em Proceedings of the IEEE international conference on computer
  vision}, pages 764--773, 2017.

\bibitem{graham20183d}
Benjamin Graham, Martin Engelcke, and Laurens van~der Maaten.
\newblock 3d semantic segmentation with submanifold sparse convolutional
  networks.
\newblock In {\em Proceedings of the IEEE Conference on Computer Vision and
  Pattern Recognition}, pages 9224--9232, 2018.

\bibitem{hua2018pointwise}
Binh-Son Hua, Minh-Khoi Tran, and Sai-Kit Yeung.
\newblock Pointwise convolutional neural networks.
\newblock In {\em Proceedings of the IEEE Conference on Computer Vision and
  Pattern Recognition}, pages 984--993, 2018.

\bibitem{huang2018recurrent}
Qiangui Huang, Weiyue Wang, and Ulrich Neumann.
\newblock Recurrent slice networks for 3d segmentation of point clouds.
\newblock In {\em Proceedings of the IEEE Conference on Computer Vision and
  Pattern Recognition}, pages 2626--2635, 2018.

\bibitem{ioffe2015batch}
Sergey Ioffe and Christian Szegedy.
\newblock Batch normalization: Accelerating deep network training by reducing
  internal covariate shift.
\newblock {\em arXiv preprint arXiv:1502.03167}, 2015.

\bibitem{jiang2018pointsift}
Mingyang Jiang, Yiran Wu, and Cewu Lu.
\newblock Pointsift: A sift-like network module for 3d point cloud semantic
  segmentation.
\newblock {\em arXiv preprint arXiv:1807.00652}, 2018.

\bibitem{kanezaki2018rotationnet}
Asako Kanezaki, Yasuyuki Matsushita, and Yoshifumi Nishida.
\newblock Rotationnet: Joint object categorization and pose estimation using
  multiviews from unsupervised viewpoints.
\newblock In {\em Proceedings of the IEEE Conference on Computer Vision and
  Pattern Recognition}, pages 5010--5019, 2018.

\bibitem{klokov2017escape}
Roman Klokov and Victor Lempitsky.
\newblock Escape from cells: Deep kd-networks for the recognition of 3d point
  cloud models.
\newblock In {\em Proceedings of the IEEE International Conference on Computer
  Vision}, pages 863--872, 2017.

\bibitem{krizhevsky2012imagenet}
Alex Krizhevsky, Ilya Sutskever, and Geoffrey~E Hinton.
\newblock Imagenet classification with deep convolutional neural networks.
\newblock In {\em Advances in neural information processing systems}, pages
  1097--1105, 2012.

\bibitem{landrieu2018large}
Loic Landrieu and Martin Simonovsky.
\newblock Large-scale point cloud semantic segmentation with superpoint graphs.
\newblock In {\em Proceedings of the IEEE Conference on Computer Vision and
  Pattern Recognition}, pages 4558--4567, 2018.

\bibitem{li2018so}
Jiaxin Li, Ben~M Chen, and Gim Hee~Lee.
\newblock So-net: Self-organizing network for point cloud analysis.
\newblock In {\em Proceedings of the IEEE conference on computer vision and
  pattern recognition}, pages 9397--9406, 2018.

\bibitem{li2018pointcnn}
Yangyan Li, Rui Bu, Mingchao Sun, Wei Wu, Xinhan Di, and Baoquan Chen.
\newblock Pointcnn: Convolution on x-transformed points.
\newblock In {\em Advances in Neural Information Processing Systems}, pages
  828--838, 2018.

\bibitem{li2016fpnn}
Yangyan Li, Soeren Pirk, Hao Su, Charles~R Qi, and Leonidas~J Guibas.
\newblock Fpnn: Field probing neural networks for 3d data.
\newblock In {\em Advances in Neural Information Processing Systems}, pages
  307--315, 2016.

\bibitem{li2015gated}
Yujia Li, Daniel Tarlow, Marc Brockschmidt, and Richard Zemel.
\newblock Gated graph sequence neural networks.
\newblock {\em arXiv preprint arXiv:1511.05493}, 2015.

\bibitem{liu2018point2sequence}
Xinhai Liu, Zhizhong Han, Yu-Shen Liu, and Matthias Zwicker.
\newblock Point2sequence: Learning the shape representation of 3d point clouds
  with an attention-based sequence to sequence network.
\newblock {\em arXiv preprint arXiv:1811.02565}, 2018.

\bibitem{maturana2015voxnet}
Daniel Maturana and Sebastian Scherer.
\newblock Voxnet: A 3d convolutional neural network for real-time object
  recognition.
\newblock In {\em 2015 IEEE/RSJ International Conference on Intelligent Robots
  and Systems (IROS)}, pages 922--928. IEEE, 2015.

\bibitem{qi2017pointnet}
Charles~R Qi, Hao Su, Kaichun Mo, and Leonidas~J Guibas.
\newblock Pointnet: Deep learning on point sets for 3d classification and
  segmentation.
\newblock In {\em Proceedings of the IEEE Conference on Computer Vision and
  Pattern Recognition}, pages 652--660, 2017.

\bibitem{qi2016volumetric}
Charles~R Qi, Hao Su, Matthias Nie{\ss}ner, Angela Dai, Mengyuan Yan, and
  Leonidas~J Guibas.
\newblock Volumetric and multi-view cnns for object classification on 3d data.
\newblock In {\em Proceedings of the IEEE conference on computer vision and
  pattern recognition}, pages 5648--5656, 2016.

\bibitem{qi2017pointnet++}
Charles~Ruizhongtai Qi, Li Yi, Hao Su, and Leonidas~J Guibas.
\newblock Pointnet++: Deep hierarchical feature learning on point sets in a
  metric space.
\newblock In {\em Advances in Neural Information Processing Systems}, pages
  5099--5108, 2017.

\bibitem{qi20173d}
Xiaojuan Qi, Renjie Liao, Jiaya Jia, Sanja Fidler, and Raquel Urtasun.
\newblock 3d graph neural networks for rgbd semantic segmentation.
\newblock In {\em Proceedings of the IEEE International Conference on Computer
  Vision}, pages 5199--5208, 2017.

\bibitem{rethage2018fully}
Dario Rethage, Johanna Wald, Jurgen Sturm, Nassir Navab, and Federico Tombari.
\newblock Fully-convolutional point networks for large-scale point clouds.
\newblock In {\em Proceedings of the European Conference on Computer Vision
  (ECCV)}, pages 596--611, 2018.

\bibitem{riegler2017octnet}
Gernot Riegler, Ali Osman~Ulusoy, and Andreas Geiger.
\newblock Octnet: Learning deep 3d representations at high resolutions.
\newblock In {\em Proceedings of the IEEE Conference on Computer Vision and
  Pattern Recognition}, pages 3577--3586, 2017.

\bibitem{ronneberger2015u}
Olaf Ronneberger, Philipp Fischer, and Thomas Brox.
\newblock U-net: Convolutional networks for biomedical image segmentation.
\newblock In {\em International Conference on Medical image computing and
  computer-assisted intervention}, pages 234--241. Springer, 2015.

\bibitem{rusu2008towards}
Radu~Bogdan Rusu, Zoltan~Csaba Marton, Nico Blodow, Mihai Dolha, and Michael
  Beetz.
\newblock Towards 3d point cloud based object maps for household environments.
\newblock {\em Robotics and Autonomous Systems}, 56(11):927--941, 2008.

\bibitem{scarselli2009graph}
Franco Scarselli, Marco Gori, Ah~Chung Tsoi, Markus Hagenbuchner, and Gabriele
  Monfardini.
\newblock The graph neural network model.
\newblock {\em IEEE Transactions on Neural Networks}, 20(1):61--80, 2009.

\bibitem{shen2018mining}
Yiru Shen, Chen Feng, Yaoqing Yang, and Dong Tian.
\newblock Mining point cloud local structures by kernel correlation and graph
  pooling.
\newblock In {\em Proceedings of the IEEE conference on computer vision and
  pattern recognition}, pages 4548--4557, 2018.

\bibitem{simonovsky2017dynamic}
Martin Simonovsky and Nikos Komodakis.
\newblock Dynamic edge-conditioned filters in convolutional neural networks on
  graphs.
\newblock In {\em Proceedings of the IEEE Conference on Computer Vision and
  Pattern Recognition}, pages 3693--3702, 2017.

\bibitem{su2018splatnet}
Hang Su, Varun Jampani, Deqing Sun, Subhransu Maji, Evangelos Kalogerakis,
  Ming-Hsuan Yang, and Jan Kautz.
\newblock Splatnet: Sparse lattice networks for point cloud processing.
\newblock In {\em Proceedings of the IEEE Conference on Computer Vision and
  Pattern Recognition}, pages 2530--2539, 2018.

\bibitem{su2015multi}
Hang Su, Subhransu Maji, Evangelos Kalogerakis, and Erik Learned-Miller.
\newblock Multi-view convolutional neural networks for 3d shape recognition.
\newblock In {\em Proceedings of the IEEE international conference on computer
  vision}, pages 945--953, 2015.

\bibitem{szegedy2015going}
Christian Szegedy, Wei Liu, Yangqing Jia, Pierre Sermanet, Scott Reed, Dragomir
  Anguelov, Dumitru Erhan, Vincent Vanhoucke, and Andrew Rabinovich.
\newblock Going deeper with convolutions.
\newblock In {\em Proceedings of the IEEE conference on computer vision and
  pattern recognition}, pages 1--9, 2015.

\bibitem{tatarchenko2018tangent}
Maxim Tatarchenko, Jaesik Park, Vladlen Koltun, and Qian-Yi Zhou.
\newblock Tangent convolutions for dense prediction in 3d.
\newblock In {\em Proceedings of the IEEE Conference on Computer Vision and
  Pattern Recognition}, pages 3887--3896, 2018.

\bibitem{wang2017dominant}
Chu Wang, Marcello Pelillo, and Kaleem Siddiqi.
\newblock Dominant set clustering and pooling for multi-view 3d object
  recognition.
\newblock In {\em Proceedings of British Machine Vision Conference (BMVC)},
  volume~12, 2017.

\bibitem{wang2018local}
Chu Wang, Babak Samari, and Kaleem Siddiqi.
\newblock Local spectral graph convolution for point set feature learning.
\newblock In {\em Proceedings of the European Conference on Computer Vision
  (ECCV)}, pages 52--66, 2018.

\bibitem{wang2015voting}
Dominic~Zeng Wang and Ingmar Posner.
\newblock Voting for voting in online point cloud object detection.
\newblock In {\em Robotics: Science and Systems}, volume~1, pages 10--15607,
  2015.

\bibitem{wang2018deep}
Shenlong Wang, Simon Suo, Wei-Chiu Ma, Andrei Pokrovsky, and Raquel Urtasun.
\newblock Deep parametric continuous convolutional neural networks.
\newblock In {\em Proceedings of the IEEE Conference on Computer Vision and
  Pattern Recognition}, pages 2589--2597, 2018.

\bibitem{wang2018sgpn}
Weiyue Wang, Ronald Yu, Qiangui Huang, and Ulrich Neumann.
\newblock Sgpn: Similarity group proposal network for 3d point cloud instance
  segmentation.
\newblock In {\em Proceedings of the IEEE Conference on Computer Vision and
  Pattern Recognition}, pages 2569--2578, 2018.

\bibitem{wang2018dynamic}
Yue Wang, Yongbin Sun, Ziwei Liu, Sanjay~E Sarma, Michael~M Bronstein, and
  Justin~M Solomon.
\newblock Dynamic graph cnn for learning on point clouds.
\newblock {\em arXiv preprint arXiv:1801.07829}, 2018.

\bibitem{wu2018pointconv}
Wenxuan Wu, Zhongang Qi, and Li Fuxin.
\newblock Pointconv: Deep convolutional networks on 3d point clouds.
\newblock {\em arXiv preprint arXiv:1811.07246}, 2018.

\bibitem{xie2018attentional}
Saining Xie, Sainan Liu, Zeyu Chen, and Zhuowen Tu.
\newblock Attentional shapecontextnet for point cloud recognition.
\newblock In {\em Proceedings of the IEEE Conference on Computer Vision and
  Pattern Recognition}, pages 4606--4615, 2018.

\bibitem{xu2015empirical}
Bing Xu, Naiyan Wang, Tianqi Chen, and Mu Li.
\newblock Empirical evaluation of rectified activations in convolutional
  network.
\newblock {\em arXiv preprint arXiv:1505.00853}, 2015.

\bibitem{xu2018spidercnn}
Yifan Xu, Tianqi Fan, Mingye Xu, Long Zeng, and Yu Qiao.
\newblock Spidercnn: Deep learning on point sets with parameterized
  convolutional filters.
\newblock In {\em Proceedings of the European Conference on Computer Vision
  (ECCV)}, pages 87--102, 2018.

\bibitem{yi2016scalable}
Li Yi, Vladimir~G Kim, Duygu Ceylan, I Shen, Mengyan Yan, Hao Su, Cewu Lu,
  Qixing Huang, Alla Sheffer, Leonidas Guibas, et~al.
\newblock A scalable active framework for region annotation in 3d shape
  collections.
\newblock {\em ACM Transactions on Graphics (TOG)}, 35(6):210, 2016.

\bibitem{yi2017syncspeccnn}
Li Yi, Hao Su, Xingwen Guo, and Leonidas~J Guibas.
\newblock Syncspeccnn: Synchronized spectral cnn for 3d shape segmentation.
\newblock In {\em Proceedings of the IEEE Conference on Computer Vision and
  Pattern Recognition}, pages 2282--2290, 2017.

\end{thebibliography}
}

\end{document}